\documentclass[letterpaper]{article} 
\usepackage{aaai17}  
\usepackage{graphicx}
\usepackage{amsmath}
\usepackage{amsfonts}
\usepackage{url}
\usepackage{algorithm, algpseudocode}
\usepackage{comment}
\usepackage{makeidx}
\usepackage{caption}
\usepackage{subcaption}
\algblock{ParForThread}{EndParForThread}
\algnewcommand\algorithmicpardo{\textbf{do}}
\algnewcommand\algorithmicparforthread{\textbf{parallelForByThreads}}
\algnewcommand\algorithmicendparforthread{\textbf{end\ parallelForByThreads}}
\algrenewtext{ParForThread}[1]{\algorithmicparforthread\ #1\ \algorithmicpardo}
\algrenewtext{EndParForThread}{\algorithmicendparforthread}

\algblock{ParForBlock}{EndParForBlock}
\algnewcommand\algorithmicparforblock{\textbf{parallelForByBlocks}}
\algnewcommand\algorithmicendparforblock{\textbf{end\ parallelForByBlocks}}
\algrenewtext{ParForBlock}[1]{\algorithmicparforblock\ #1\ \algorithmicpardo}
\algrenewtext{EndParForBlock}{\algorithmicendparforblock}

\algtext*{EndWhile}
\algtext*{EndFor}
\algtext*{EndIf}
\algtext*{EndFunction}
\algrenewcommand\algorithmicindent{0.5em}
\usepackage{varwidth}

\newif\ifextended 
\extendedtrue 

\begin{document}

\title{Block-Parallel IDA* for GPUs
\ifextended
\\ (Extended Manuscript)
\fi
}
 \author{Satoru Horie \and Alex Fukunaga\\ 
Graduate School of Arts and Sciences\\
The University of Tokyo
}

\maketitle

\begin{abstract}
We investigate GPU-based parallelization of Iterative-Deepening A* (IDA*).
We show that straightforward thread-based parallelization techniques which were previously proposed for massively parallel SIMD processors perform poorly due to warp divergence and load imbalance.
We propose Block-Parallel IDA* (BPIDA*), which assigns the search of a subtree  to a block (a group of threads with access to  fast shared memory) rather than a thread.
On the 15-puzzle, BPIDA* on a NVIDIA GRID K520 with 1536 CUDA cores achieves  a speedup of 4.98 compared to a highly optimized sequential IDA* implementation on a Xeon E5-2670 core.
\ifextended \footnote{This is an extended manuscript based on a paper accepted to appear in SoCS2017.}

\end{abstract}

\section{Introduction}

Graphical Processing Units (GPUs) are many-core processors which 
are now widely used to accelerate many types of computation.
GPUs are attractive for combinatorial search because of their massive parallelism. On the other hand, on  many domains, search algorithms such as A* tend to be limited by RAM rather than runtime. 
A standard strategy for addressing limited memory in sequential  search is iterative deepening \cite{korf85}. 
We present a case study on  the GPU-parallelization of Iterative-Deepening A* \cite{korf85}
for the 15-puzzle using the Manhattan Distance heuristic.
We evaluate previous thread-based techniques for parallelizing IDA* on SIMD machines, and show that these do not scale well due to poor load balance and warp divergence. 
We then propose Block-Parallel IDA* (BPIDA*), 
which, instead of assigning a subtree to a single thread, assigns a subtree to a group of threads which share fast memory. 
BPIDA*  achieves a speedup of 4.98 compared to a state-of-the-art 15-puzzle solver on a CPU, and a speedup of 659.5 compared to a single-thread version of the code running on the GPU.




\section{Background and Related Work}

An NVIDIA CUDA architecture  GPU  consists of a set of {\em streaming multiprocessors (SMs)} and a GPU main memory (shared among all SMs).
Each SM consists of shared memory, cache, registers, arithmetic units, and a warp scheduler.
Within each SM the cores operate in a SIMD manner.
However, each SM executes independently, so threads in different SMs can run asynchronously.
A {\em thread} is the smallest unit of execution.
A {\em block} is a group of threads which execute on the same SM and share memory.
A {\em grid} is a group of blocks which execute the same function.
%
Threads in a block are partitioned into {\em warps}. A warp executes in a SIMD manner (all threads in the same warp share a program counter).
{\em Warp divergence}, an instance of {\em SIMD divergence}, occurs when threads belonging to the same warp follow different execution paths, 
e.g., IF-THEN-ELSE branches.
%
{\em Shared memory} is shared by a block and is local within a SM, and access to shared memory is much faster than access to the GPU {\em global memory} which is shared by all SMs.

Rao et al (\citeyear{rao87}) parallelized each iteration of IDA* 
 using work-stealing on multiprocessors. 
Parallel-window IDA* assigned each iteration of IDA* to its own processor \cite{powley89}. 
Two SIMD parallel IDA* algorithms are by Powley et al 
(\citeyear{powley93}) and \citeauthor{mahanti93} (\citeyear{mahanti93}).
For each $f$-cost limited iteration of IDA*, they perform an initial partition of the workload among the processors, and then periodically perform load balancing between IDA* iterations and within each iteration. 
Hayakawa et al 
(\citeyear{hayakawa15}) proposed a GPU-based parallelization of IDA* for the 3x3x3 Rubik's cube which
searches to a fixed depth $l$ on the CPU, then invokes a GPU kernel for the remaining subproblems. 
Their domain-specific load balancing scheme relies on tuning $l$ using knowledge of ``God's number'' (optimal path length for the most difficult cube instance) and is fragile --  perturbing $l$ by 1 results in 
a 10x slowdown. 
%
\citeauthor{zhou15} (\citeyear{zhou15}) proposed a GPU-parallel A* which partitions OPEN into thousands of priority queues. 
The amount of global RAM  on the GPU (currently $\leq 24GB$) poses a serious limitation for GPU-based parallel A*.
Edelkamp and Sulewski (\citeyear{edelkamp2010perfect}) investigated memory-efficient GPU search.
Sulewski et al (\citeyear{sulewski2011exploiting}) proposed a hybrid planner which uses both the GPU and CPU.

\section{Experimental Settings and Baselines}
\label{sec:baseline}

We used the standard set of 100 15-puzzle instances by Korf (\citeyear{korf85}).
These instances are ordered in approximate order of difficulty.
All solvers used the Manhattan distance heuristic.
Reported runtimes include all overheads such as data transfers between CPU and GPU memories   
(negligible).
All experiments were executed on a non-shared, dedicated AWS EC2
g2.2xlarge instance.
The CPU is an Intel Xeon E5-2670.
The GPU is an NVIDIA GRID K520, with
4GiB global RAM, 
48KiB  shared RAM/block,
1536 CUDA cores, 
warp size 32, and
0.80GHz GPU clock rate.

First, we evaluated 3 baseline IDA* solvers:\\
\noindent {\bf Solver B}: The efficient, Manhattan-Distance heuristic based  15-puzzle IDA* solver implemented in C++ by \citeauthor{burns12} (\citeyear{burns12}). We used the current version at https://github.com/eaburns/ssearch.\\
\noindent {\bf Solver C}: Our own implementation of IDA* in C (code at http://github.com/socs2017-48/anon48), 
This is the basis for G1 and all of our GPU-based code.\\
\noindent {\bf Solver G1}: A direct port of Solver C to CUDA. The implementation is optimized so that all data structures are in the fast, shared memory (the memory which is local to a SM). {\bf This baseline configuration uses only 1 GPU block/thread, i.e., only 1 core is used, all other GPU cores are idle}.

The total time to solve all 100 problem instances was 620 seconds for Solver B \cite{burns12} and 475 seconds for our Solver C.
Solver C was consistently ~25\% faster on every instance.
Thus, Solver C is appropriate as a baseline for our GPU-based 15-puzzle solvers.

Next, we compare Solver C (1 CPU thread) to G1 (1 GPU thread).
G1 required 62957 seconds to solve all 100 instances, 131 times slower than Solver C.
This implies that on the GPU we used with 1536 cores, a perfectly efficient implementation of parallel IDA* might be able to achieve a speedup of  up to 1536/131 = 11.725 compared to Solver C.

\section{Thread-Based  Parallel IDA*}

Most of the previous work on parallel IDA* parallelizes each iteration of IDA* using a {\em thread-based parallel} scheme 
\cite{rao87,powley93,mahanti93,hayakawa15}.

We evaluated 3 thread-parallel IDA* configurations. 
Since these are relatively straightforward and not novel, we sketch the implementations below. 
\ifextended
  Details are in Appendix.
\else
  Details are in the extended version \cite{ExtendedArxivManuscript}.
\fi 

\subsubsection*{PSimple (baseline)}
In this baseline configuration,  for each $f$-bounded iteration of IDA*, 
PSimple performs A* search from the start state until as many unique states as the \# of threads are in OPEN. Then, each root is assigned to a thread.
No load balancing is performed. 
The subtree sizes under each root state can vary significantly, so some threads may finish their subproblem much faster than other threads.
Each $f$-bounded iteration must wait for all threads to complete, so PSimple has very poor load balance.
Therefore, {\em load balancing} mechanisms which redistribute the work among processors are necessary. 

\subsubsection*{PStaticLB (static load balancing)}
\label{sec:static-load-balancing}
This configuration adds static load balancing to PSimple.
After each $f$-bounded iteration, PStaticLB implements a  {\em static load balancing} mechanism somewhat similar to that of \cite{powley93}.
In IDA*, the $i$-th iteration repeats all of the work done in iteration $i-1$. 
Thus, the \# of states visited under each root state in the iteration $i-1$ can be used to estimate the \# of states which will be visited in the current iteration $i$, and root nodes are redistributed based on these estimates 
\ifextended
 (details in Appendix).
\else
 (details in \cite{ExtendedArxivManuscript}).
\fi

\subsubsection*{PFullLB (thread-parallel with dynamic load balancing)}
\label{sec:dynamic-load-balancing}
This configuration adds dynamic load balancing (DLB) to PStaticLB, 
which  moves work to idle threads from threads with remaining work {\em during} an iteration.
On a GPU, work can be transferred between two threads {\em within} a single block relatively cheaply using the shared memory within a block, while  transferring work between two threads in different blocks  is expensive because it requires access to the global memory.
When dynamic load balancing is triggered, idle threads steal work from threads with remaining work within a block.
We experimented with various DLB strategies including variants of policies investigated by \cite{powley89,mahanti93}, and 
used a  policy we found for triggering DLB  
based on the policy by \citeauthor{powley89}. 
\ifextended
See Appendix for additional details.
\else
See \cite{ExtendedArxivManuscript} for additional details.
\fi

\subsection{Evaluation of Thread-Parallel IDA*}

PSimple on 1536 cores required a total of 3378 seconds to solve all 100 problems, a speedup of only 18.6 compared to G1 (1 core on the GPU).
This is mostly due to extremely poor load balance.
We define load balance as $\mathit{maxload}/\mathit{averageload}$, where $\mathit{averageload}$ is the average number of nodes expanded among all threads, and $\mathit{maxload}$ is the number of states expanded by the thread which performed the most work. 
The load balance for PSimple on the 100 problems was: mean 96.46, min 14, max 680, stddev 113.19.
This is extremely unbalanced ($maxload$ is almost 100x $averageload$).

Static load balancing significantly improved load balance (PStaticLB: mean 9.96, min 3, max 56, stddev 8.96),
and dynamic load balancing further improved load balance (PFullLB: mean 6.14 min 3 max 19 stddev 3.38).
This resulted in speedups of 58.9 and 70.8 compared to G1 (Table \ref{tab:summary-runtime}).
However, the 70.8 speedup vs G1 achieved by PFullLB is only a parallel efficiency of 70.8/1536 = 4.6\%, which is extremely poor.
We experimented extensively but could not achieve significantly better results with thread-parallel IDA*.

\section{Block Parallelization}
\label{sec:BlockParallelization}
The likely causes for  the poor  (4.6\%) efficiency of PFullLB are:
(1) SMs become idle due to poor load balance even after our load balancing efforts,
(2) threads stall for warp divergence, and
(3) load balancing overhead.
All of these can be attributed to the thread-based parallelization scheme in PFullLB and PStaticLB, in which each processor/thread executes an independent subproblem during a single $f$-bound iteration.
This scheme, based on parallel IDA* variants originally designed for SIMD machines \cite{powley93,mahanti93}, was appropriate for those SIMD architectures where all communications between processors were very expensive -- paying the price of SIMD divergence overhead was preferable to incurring communication costs.
On the other hand, in NVIDIA GPUs, threads in the same block (which execute on the same SM) can access fast shared memory on the SM with relatively low overhead.
We exploit this in a {\em block-parallel} approach.








\citeauthor{rocki09} (\citeyear{rocki09}) proposed a GPU-based parallel minimax gametree search algorithm for 8x8 Othello (without any $\alpha\beta$ pruning) which works as follows.
Within each block a node $n$ is selected for expansion. If $n$ is a leaf, it is evaluated using a parallel evaluation function (32 threads, 1 thread per 2 positions in the 8x8 board). Otherwise a parallel successor generator function is  called (1 thread/position) to generate successors of $n$, which are added to the node queue.
This  approach greatly reduced warp divergence, since all threads in the warp are synchronized to perform the fetch-evaluate-expand cycle.
Because there is no $\alpha\beta$ pruning, their search trees have uniform depth (i.e., fixed-depth DFS), and also, the \# of possible moves on the othello board (64) conveniently matched a multiple of the CUDA warp size (32).

We now propose a generalization of this approach for IDA*, which we call {\em Block-Parallel IDA*}, shown in Alg.~\ref{alg:Block-Parallel-IDA*}.
In contrast to the parallel minimax of \cite{rocki09}, BPIDA* handles variable-depth subtrees  (due to the heuristic, IDA* tree depths are irregular) and does not depend on a fixed number of applicable operators (e.g., 64).

$openList$ is a stack which is shared among all threads in the same block, which
supports two key parallel operations: {\tt parallelPop} and {\tt atomicPut}.
{\tt parallelPop} extracts $(\mathit{\#threads\_in\_a\_block} / \mathit{\#operators})$ nodes from $openList$.
{\tt atomicPut} inserts nodes in $t$ into the shared $openList$ concurrently. This is implemented as a linearizable~\cite{herlihy90} operation. 

The {\tt BPDFS} function is similar to a standard, sequential $f$-limited depth-first search, 
but in each iteration of the repeat-until loop in lines 4-16 (Alg.~\ref{alg:Block-Parallel-IDA*}), 
a warp performs the fetch-evaluate-expand cycle on $(\mathit{\#threads\_in\_a\_block} / \mathit{\#operators})$ nodes.
The number of threads per block is set to the warp size (32). This allows the following:
(1) When a warp is scheduled for execution, all cores in the SM are active.
(2) Since all threads in the block (=warp) share a program counter, explicit synchronizations become unnecessary.

BPIDA* applies a slightly modified version of the static load balancing  used by PStaticLB
(Sec.~\ref{sec:static-load-balancing}).
While PStaticLB uses the number of expanded nodes 
to estimate the work in the next iteration,
BPIDA* uses the number of repetitions executed in lines 4-16. 
BPIDA* does not use dynamic load balancing.

\begin{algorithm}[tb]
	\caption{BlockParallel IDA*}
	\label{alg:Block-Parallel-IDA*}
\begin{algorithmic}[1]
\footnotesize
	\Function{BPDFS}{$root, goals, limit_f$}
	\State $openList \gets root$
	\State $f_{next} = \infty$
	\Repeat
		\State $s \gets \Call{parallelPop}{openList}$
		\If{$s \in goals$}
			\State \Return $s$ and its parents as a shortest path
		\EndIf
		\State $a \gets (\mathit{threadID} \enspace \text{mod} \enspace \text{\#actions}) \text{th action}$
		\If{$a$ is applicable on $s$}
			\State $t \gets successor(a, s)$
			\State $f_{new} \gets g(s) + cost(a) + h(t)$
			\If{$f_{new} <= limit_f$}
				\State $\Call{atomicPut}{openList, t}$
			\Else
				\State $f_{next} \gets min(f_{next}, f_{new})$
			\EndIf
		\EndIf
	\Until{$openList$ is empty}
	\State \Return $f_{next}$ \Comment no plan is found 
	\EndFunction
\\
	\Function{BPIDA*}{$start, goals$}
		\State $roots \gets \Call{CreateRootSet}{start, goals}$
		\State $limit_f \gets \Call{DecideFirstLimit}{roots}$
		\Repeat
			\ParForBlock{$root \in roots$}
				\State $limit_f, stat \gets \Call{BPDFS}{root, goals, limit_f}$
			\EndParForBlock
		\Until{shortest path is found}
	\EndFunction
\end{algorithmic}
\end{algorithm}

\section{Evaluation of BPIDA*}

\begin{figure*}[htb]
\centering
\begin{subfigure}[b]{.31\textwidth}
\includegraphics[width=\linewidth]{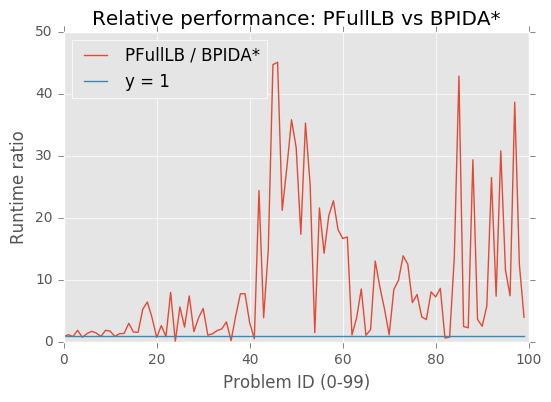}
\caption{Relative runtimes: PFullLB vs. BPIDA*}
\label{block_rel}
\end{subfigure}
\begin{subfigure}[b]{.31\textwidth}
\includegraphics[width=\linewidth]{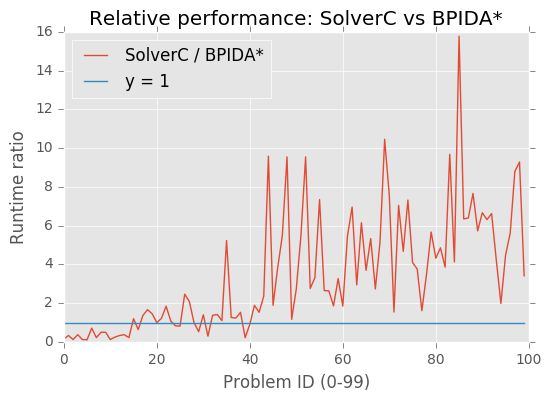}
\caption{Relative runtimes: Solver C vs. BPIDA* (finding 1 optimal solution)}
\label{one_rel}
\end{subfigure}
\begin{subfigure}[b]{.31\textwidth}
\includegraphics[width=\linewidth]{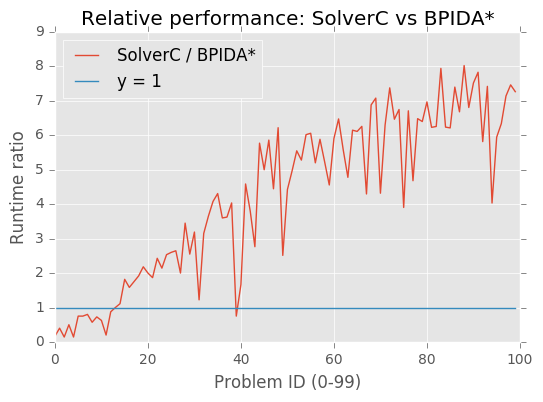}
\caption{Relative runtimes: Solver C vs. BPIDA* (finding all optimal solutions)}
\label{allsol_rel}
\end{subfigure}
\caption{BP-IDA* Evaluation}
\end{figure*}

\subsubsection*{Runtimes}

Figure \ref{block_rel} compares the relative runtime of BPIDA* vs. PFullLB.
BPIDA* required a total of 95 seconds to solve all 100 problems, a speedup of 9.39  compared to PFullLB.
Table \ref{tab:summary-runtime} summarizes the total runtimes and speedups for all algorithms in this paper.

\setlength{\textfloatsep}{0.2cm}
\captionsetup{skip=0pt}
\begin{table}[b]
  \begin{center}
\footnotesize
  \begin{tabular}{|l|cc|} \hline
  configuration  &  total runtime  &  speedup \\
                 & (seconds) & vs. G1 \\
                 \hline
\multicolumn{3}{|c|}{CPU-based sequential algorithms (1 CPU thread)}\\
  \hline
  Solver B \cite{burns12}   & 620 & n/a\\
  Solver C & 475 & n/a\\
  \hline
\multicolumn{3}{|c|}{GPU-based sequential algorithm (1 thread)}\\
  \hline
  G1  &   62957 & 1 \\
  \hline
\multicolumn{3}{|c|}{GPU-based parallel algorithms (1536 threads)}\\
  \hline
  PSimple &  3378& 18.6\\
  PStaticLB & 1069 & 58.9\\
  PFullLB & 892&  70.8 \\
  {\bf BPIDA*} & {\bf 95} & {\bf 659.5}\\ \hline
  \end{tabular}
  \end{center}
  \caption{Total Runtimes for 100 15-Puzzle Instances}
  \label{tab:summary-runtime}
\end{table}

\subsubsection*{Other metrics}
There are 3 suspected culprits for the poor performance of thread-based parallel IDA*: (1) dynamic load overhead, (2) idle SMs (bad load balance), and (3) thread stalls for warp divergence.
BPIDA* doesn't perform dynamic load balancing, so (1) is irrelevant.
For factors (2) and (3), there are related metrics, sm\_efficiency and IPC(instructions per cycle), which can be measured by the CUDA profiler, nvprof.
sm\_efficiency is the average \% of time at least one warp is active on a SM.
High sm\_efficiency shows how busy the SMs are, and high IPC indicates there are few NOPs due to warp divergence.
The  (mean, min, max, stddev) sm\_efficiency over 100 instances  was (65.22, 31.8, 82.7, 7.94) for PFullLB, and (94.29, 32.3, 99.9, 9.76) for BPIDA*, and 
for  IPC, the results were (0.30, 0.13, 0.39, 0.048) for PFullLB and  (0.97, 0.60, 1.06, 0.059) for BPIDA*.
For both metrics, the results of BPIDA* were better than PFullLB, and close to the ideal values (100\% sm\_efficiency and IPC=1.0).

\subsection{Comparison with Sequential Solver C}

We now compare BPIDA* with the CPU-based, sequential Solver C (Sec.~\ref{sec:baseline}).
Fig. \ref{one_rel} compares the relative runtimes of Solver C (1 CPU core) and BPIDA* (1536 GPU cores). The y-axis shows Runtime(SolverC)/Runtime(BPIDA*) for each instance.
Comparing the total time to solve all 100 instances, BPIDA* was 4.98 times faster.

Runtime comparisons between  parallel vs. sequential IDA* can be obfuscated by the fact that 
they do not necessarily expand the same set of nodes in the final iteration, although they expand the same set of nodes in non-final iterations (the same issue exists with comparisons among parallel IDA* variants, but from Fig. \ref{block_rel} and  Table \ref{tab:summary-runtime}, it is clear that BPIDA* significantly outperforms the other parallel algorithms, so above, we simply reported the time to find a single solution, as is standard practice in previous works).

To eliminate differences in search efficiency (node expansion order) from the comparison, the next experiment compares  the time required to find {\em all optimal-cost solutions} of every problem, i.e., the search does not terminate until all nodes with $f \leq OptimalCost$ have been expanded.
This eliminates node ordering effects, allowing comparison of the wall-clock time required to perform the same amount of search.
Fig. \ref{allsol_rel} compares the relative runtimes of Solver C (1 CPU core) and BPIDA* (1536 GPU cores). The y-axis shows Runtime(SolverC)/Runtime(BPIDA*) to find all optimal solutions for each instance.
Comparing the total time to find all optimal solutions for all 100 instances, BPIDA* was 6.78 times faster.

\section{Conclusions and Future Work}
We proposed Block-Parallel IDA*, which assigns subtrees to GPU blocks (groups of threads with fast shared memory).
Compared to thread-parallel approaches, this greatly reduces warp divergence and improves load balance. BPIDA* also does not require explicit dynamic load balancing, making it relatively simple to implement.
On 1536 cores, BPIDA* achieves a speedup of 659.5 vs. a 1-thread GPU baseline, i.e., 42\% parallel efficiency.
Compared to a highly optimized single-CPU IDA*, BPIDA* achieves a 6.78x speedup when comparing the time to find all optimal solutions.

The successful parallelization of BPIDA* on the 15-puzzle with Manhattan distance (MD) heuristic exploits the following factors:
(1) compact states, 
(2) the MD heuristic requires little memory, and
(3) standard IDA* doesn't perform duplicate state detection.
Thus,  all work could be performed in the SM local+shared memories, without using global memory. 
In many domains, data structures representing each state are larger and  the IDA* state stacks will not fit in local memory.
Also, some powerful memory-intensive heuristics, e.g,. PDBs \cite{korf02}, will require at least the use of global memory. 
Finally, standard approaches for reducing duplicate state expansion, e.g., transposition tables \cite{reinefeld94} requires significant memory. 
Thus, future work will focus on methods which use GPU global memory effectively so that domains with larger states, memory-intensive heuristics, and memory-intensive duplicate pruning techniques can be used.

\bibliographystyle{aaai}
\bibliography{main}

\ifextended


\appendix
\section{Appendices}
\subsection{Thread-Based  Parallel IDA* Details}

We provide further details on the Thread-based parallel IDA* implementations which were sketched in Section 4 of the submission.
The overall algorithmic scheme is shown in Alg. \ref{alg:Parallel-IDA*}.

\subsubsection*{PSimple (baseline)}
In this baseline configuration,  for each $f$-bounded iteration of IDA*, CreateRootSet (Alg. \ref{alg:Parallel-IDA*}, line 20) performs A* search from the start state until as many unique states as the number of threads are in OPEN. Then,  each root is assigned to a thread.
No load balancing is performed, i.e., UpdateRootSet and DynamicLoadBalance do nothing.
The subtree size under each root state can vary significantly, so some threads may finish subproblem much faster than other threads.
Each $f$-bounded iteration must wait for all threads to complete, so PSimple has very poor load balance.
Therefore, some {\em load balancing} mechanisms which (re)distribute the work among processors are necessary. 

\subsubsection*{PStaticLB (static load balancing)}
\label{sec:static-load-balancing}
This configuration adds static load balancing to PSimple.
In IDA*, the $i$-th iteration repeats all of the work done in iteration $i-1$. 
Thus, the number of states visited under each root state in the iteration $i-1$ can be used to estimate the number of states which will be visited in the current iteration $i$.

UpdateRootSet (Alg. \ref{alg:Parallel-IDA*}, line 26) implements the following {\em static load balancing} mechanism.
Let $\mathit{load(n)}$ be the number of nodes expanded in the previous iteration under node $n$.
If $\mathit{load(n)} > \mathit{averageload}$, where $\mathit{averageload}$ is the average number of nodes expanded under all of the root nodes, we {\em split} $n$ as follows.
We then initialize $\mathit{droots}$ to $\o$, and perform an A* expansion of the search tree starting at $n$, 
adding generated nodes into $\mathit{droots}$, 
until $\mathit{droots}$  has $\geq \mathit{load}(n) / \mathit{averageload}$ nodes.
Then, we remove $n$ from the root set, 
 set $load(m) \leftarrow  load(n)/|\mathit{droots}|$ for each $m \in \mathit{droots}$, 
and add $m \in \mathit{droots}$ to the root set.





Then, we allocate each root in the root set to threads.
For each thread $t$, roots are assigned to $t$  until the sum of the $\mathit{load(n)}$ for the roots assigned to $t$ exceeds
$\mathit{averageload}$.
Roots are allocated to threads in the order which they were generated -- we experimented with other orders for this assignment but did not obtain significant differences.

A goal node might be found during this split-redistribute phase, but the path to such a goal node  may or may not be optimal-cost, so it needs to be returned to OPEN without expanding it.


This rebalancing phase is performed on the CPU, so duplicate detection and pruning via the CLOSED list is performed during rebalancing.

\subsubsection*{PFullLB (thread-parallel with dynamic load balancing)}
\label{sec:dynamic-load-balancing}
This configuration adds dynamic load balancing to PStaticLB, i.e., 
the DynamicLoadBalance function (Alg. \ref{alg:Parallel-IDA*}, line 8) is enabled.

Dynamic load balancing moves work to idle threads from threads with remaining work {\em during} an iteration.
On a GPU, work can be transferred between two threads {\em within} a single block relatively cheaply because this can be done using the shared memory within a block, while  transferring work between two threads which are in different blocks  is expensive because it requires access to the global memory.
When dynamic load balancing is triggered, then within a block, idle threads steal work from threads with remaining work.

We experimented with various dynamic load balancing strategies including variants of policies investigated by \cite{powley89,mahanti93}.
The best performing policy for triggering dynamic balancing (the trigger is checked in algorithm \ref{alg:Parallel-IDA*}, line 8) is based on the policy by \citeauthor{powley89}: perform load balancing when the fraction of idle (completed) threads within a block exceeds $W(t)/(L+t)$, where $L$ is the time spent for the previous load balancing operation, $t$ is the time  since the previous load balancing operation, $W(t)$ is the amount of work (number of nodes visited) since the previous load balancing operation.
Note that time $t$ is not wall-clock time since the last rebalancing, but the actual amount of GPU time consumed.
In addition, load rebalancing is constrained so that it can not be triggered until at least $L/2$ seconds have passed since the previous rebalancing.

\begin{algorithm}[tb]
  \caption{Parallel IDA*}
  \label{alg:Parallel-IDA*}

  \begin{algorithmic}[1]
    \footnotesize
    \Function{DFS}{$root, goals, limit_f$}
    \State $openList \gets root$
    \State $f_{next} = \infty$ 
    \Repeat
    \State $s \gets \Call{pop}{openList}$
    \If{$s \in  goals$}
			\State \Return $s$ and its parents as a shortest path
    \EndIf
    \If{dynamic load balance is triggered}
    \Call{DynamicLoadBalance}{stat}
    \EndIf
    \ForAll{$a \in applicable\_actions(s)$}
	\State $t \gets successor(a, s)$
    \State $f_{new} \gets g(s) + cost(a) + h(t)$
    \If{$f_{new} <= limit_f$}
    \State $openList \gets t$
    \Else
    \State $f_{next} \gets min(f_{next}, f_{new})$
    \EndIf
    \EndFor
    \Until{$openList$ is empty}
    \State \Return $f_{next}, stat$ \Comment no plan is found
    \EndFunction
    \\
    \Function{ParallelIDA*}{start, goals}
    \State $rootset \gets \Call{CreateRootSet}{start, goals}$
    \State $limit_f \gets \Call{DecideFirstLimit}{rootSet}$
    \Repeat
    \ParForThread{$root \in rootset$}
    \State $limit_f, stat \gets \Call{DFS}{root, goals, limit_f}$
    \EndParForThread
    \State \Call{UpdateRootSet}{start, goals, rootSet, stat}
    \Until{shortest path is found}
    \EndFunction
  \end{algorithmic}
\end{algorithm}

\subsection{Evaluation of Thread-Parallel IDA*}

PSimple on 1536 cores required a total of 3378 seconds to solve all 100 problems, a speedup of only 18.6 compared to G1 (1 core on the GPU).

\begin{figure}[hbt]
\begin{center}
\includegraphics[height=5cm]{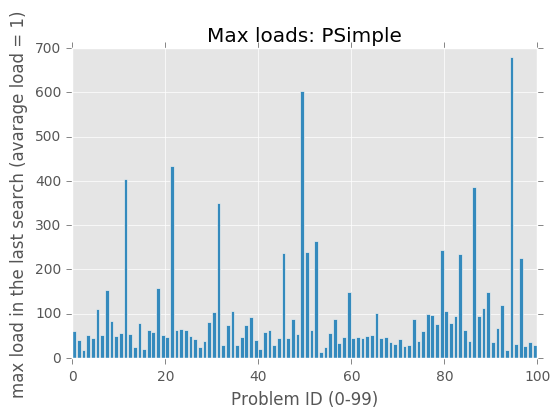}
\caption{Maximum thread workload in PSimple}
\label{baseline_loads}
\end{center}
\end{figure}

We define load balance as maxload/averageload, where averageload is the average number of nodes visited among all threads, and maxload is the number of states visited by the thread which performed the most work (visited the largest number of nodes).

Figure~\ref{baseline_loads} plots, for each of the 100 problem instances, the maximum workload (number of nodes visited) among all threads of PSimple relative to the average workload,  for the next-to-last iteration , i.e., a value of 1 means perfect load balance. We measure load balance for the next-to-last iteration because in the last iteration, only part of the tree needs to be expanded.
Figure \ref{baseline_loads} shows that some threads expand up to 600 times as many nodes as the average thread, indicating that PSimple has extremely poor load balance.

\begin{figure*}[htb]
\centering
\begin{subfigure}[b]{.31\textwidth}
\includegraphics[width=\linewidth]{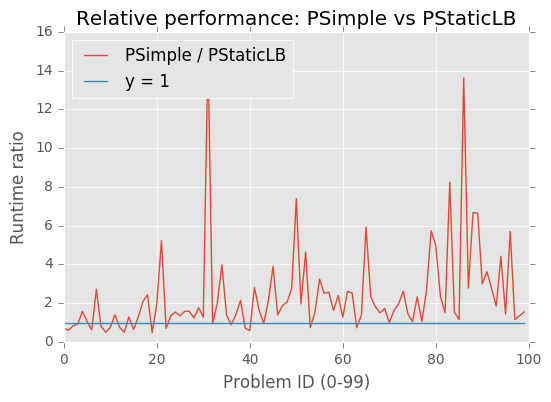}
\caption{Relative runtimes: PSimple vs. PStaticLB}
\label{static_rel}
\end{subfigure}
\begin{subfigure}[b]{.31\textwidth}
\includegraphics[width=\linewidth]{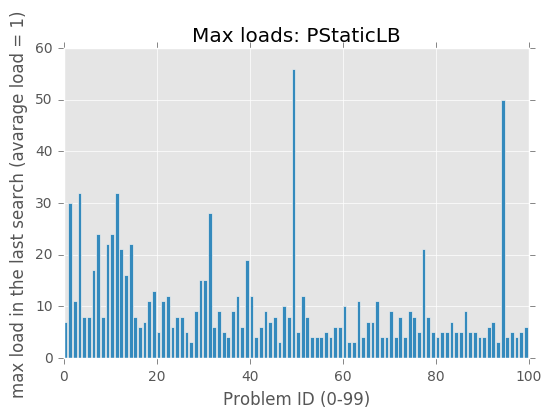}
\caption{Maximum thread workload: PStaticLB}
\label{static_loads}
\end{subfigure}
\begin{subfigure}[b]{.31\textwidth}
\includegraphics[width=\linewidth]{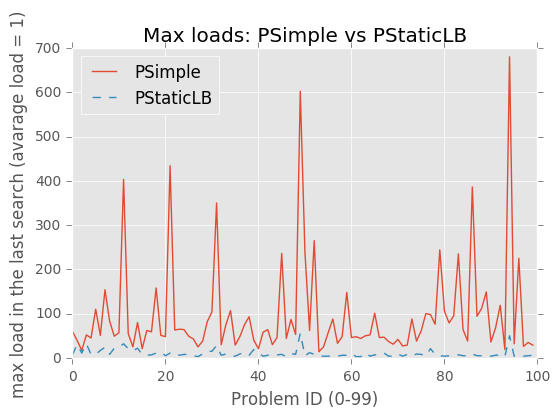}
\caption{Max workload comparison: PSimple vs. PStaticLB}
\label{static_loads_comp}
\end{subfigure}
\centering
\begin{subfigure}[b]{.31\textwidth}
\includegraphics[width=\linewidth]{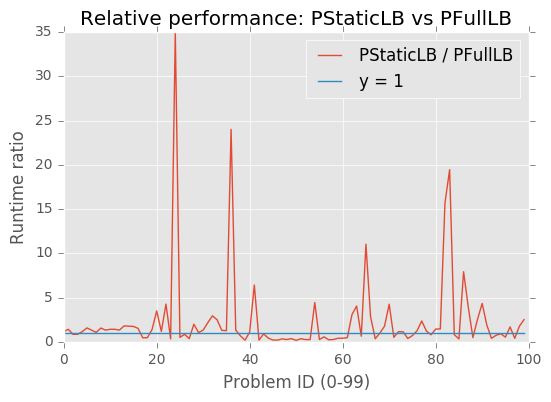}
\caption{Relative runtimes: PStaticLB vs. PFullLB}
\label{dynamic_rel}
\end{subfigure}
\begin{subfigure}[b]{.31\textwidth}
\includegraphics[width=\linewidth]{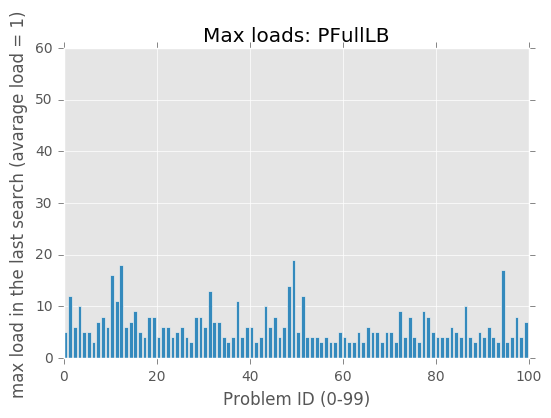}
\caption{Maximum thread Workload: PFullLB}
\label{dynamic_loads}
\end{subfigure}
\begin{subfigure}[b]{.31\textwidth}
\includegraphics[width=\linewidth]{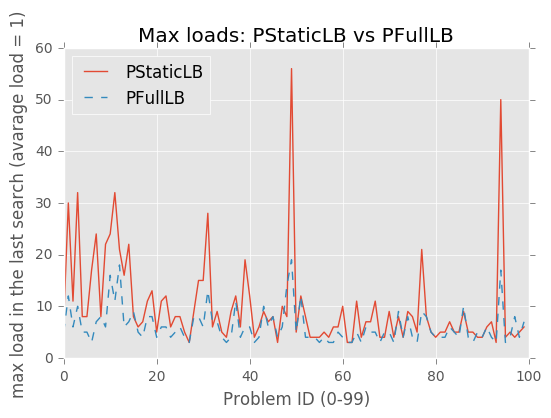}
\caption{Max workload comparison: PStaticLB vs. PFullLB}
\label{dynamic_loads_comp}
\end{subfigure}

\caption{PStaticLB Evaluation}
\end{figure*}

Figure \ref{static_rel} compares the relative runtimes, i.e., the y-axis shows Runtime(PSimple)/Runtime(PStaticLB) for each instance.
PStaticLB required a total of 1069 seconds to solve all 100 problems, a speedup of 3.16 compared to PSimple.

Figure~\ref{static_loads} plots, for each of the 100 problem instances, the maximum load (number of nodes visited) among all threads of PStaticLB relative to the average load for all threads for the next-to-last iteration.
On one hand, Figure \ref{static_loads} indicates that $maxload/averageload$ of PStaticLB is significantly smaller than that of PSimple. This explains the 3.16x speedup of PStaticLB compared to PSimple. 
On the other hand, the load balance is still quite poor, with some problems having a $maxload/averageload$ ratio $>$ 50, with most problem instances having a $maxload/averageload > 5$.

Figure \ref{dynamic_rel} compares the relative runtimes, i.e., the y-axis shows Runtime(PStaticLB)/Runtime(PFullLB) for each instance.
PFullLB required a total of 892 seconds to solve all 100 problems, a speedup of 1.20 compared to PStaticLB.

Figure~\ref{dynamic_loads} plots, for each of the 100 problem instances, the maximum load (number of nodes visited) among all threads of PFullLB relative to the average load for all threads for the next-to-last iteration.
Figure ~\ref{dynamic_loads_comp} overlays the results in Figure \ref{dynamic_loads} with the results in Figure \ref{static_loads}.
Figures \ref{dynamic_loads}-\ref{static_loads} indicate that $maxload/averageload$ of PFullLB is significantly smaller than that of PStaticLB. In particular, the worst $maxload/averageload$ ratios have been reduced to $< 20$ (compared to $>50$ for PStaticLB).
Nevertheless, most of the $maxload/averageload$ ratios are $>4$, indicating that even the combination of static and dynamic load balancing is insufficient for achieving good load balance.

We experimented extensively with both static and dynamic load balancing strategies, but so far, we have not found  any strategy/configuration that significantly outperforms the results presented here.

\fi


\end{document}